\documentclass{article}
\usepackage[utf8]{inputenc}
\usepackage{authblk}
\usepackage{setspace}
\usepackage[margin=1.25in]{geometry}
\usepackage{graphicx}
\graphicspath{ {./figures/} }
\usepackage{amsmath}

\usepackage{algorithmic}
\usepackage[boxed,commentsnumbered,linesnumbered,ruled,vlined]{algorithm2e}
\usepackage{color}
\usepackage{epsfig}
\usepackage{dsfont}
\usepackage{mathtools}
\usepackage{amssymb}
\usepackage{boldline,multirow}
\usepackage{diagbox}
\usepackage{color}
\usepackage{graphicx}
\usepackage{subfigure}
\usepackage{bbding}
\usepackage[colorlinks,
linkcolor=red,
anchorcolor=blue,
citecolor=blue
]{hyperref} 

\usepackage[style=ieee, 
citestyle=numeric-comp,
sorting=none]{biblatex}
\begin{document}

\title{Multi-site Organ Segmentation with Federated Partial Supervision and Site Adaptation}

\author[1$\dag$]{Pengbo Liu}
\author[2,3$\dag$]{Mengke Sun}
\author[1,3*]{S. Kevin Zhou}

\affil[1]{Center for Medical Imaging, Robotics, Analytic Computing \& Learning (MIRACLE), School of Biomedical Engineering \& Suzhou Institute for Advanced Research, University of Science and Technology of China, Suzhou 215123, China}
\affil[2]{Beijing UCAS Space Technology Co., Ltd., Beijing, China.}
\affil[3]{Key Lab of Intelligent Information Processing of Chinese Academy of Sciences (CAS), Institute of Computing Technology, CAS, Beijing 100190, China}
\affil[*]{Corresponding author. Email: skevinzhou@ustc.edu.cn}
\affil[$\dag$]{These authors contributed equally to this work.}

\date{}

\onehalfspacing


\maketitle

\begin{abstract}
\textit{Objective and Impact Statement:} Accurate organ segmentation is critical for many clinical applications at different clinical sites, which may have their specific application requirements that concern different organs.
\textit{Introduction:} However, learning high-quality, site-specific organ segmentation models is challenging as it often needs on-site curation of a large number of annotated images. Security concerns further complicate the matter. \textit{Methods:} The paper aims to tackle these challenges via a two-phase aggregation-then-adaptation approach. The first phase of \underline{federated aggregation} learns a single multi-organ segmentation model by leveraging the strength of `bigger data', which are formed by (i) aggregating together datasets from multiple sites that with different organ labels to provide partial supervision, and (ii) conducting partially supervised learning without data breach.
The second phase of \underline{site adaptation} is to transfer the federated multi-organ segmentation model to site-specific organ segmentation models, one model per site, in order to further improve the performance of each site's organ segmentation task.
Furthermore, improved marginal loss and exclusion loss functions are used to avoid `knowledge conflict' problem in a partially supervision mechanism. 
\textit{Results and Conclusion:} Extensive experiments on five organ segmentation datasets demonstrate the effectiveness of our multi-site approach, significantly outperforming the site-per-se learned models and achieving the performance comparable to the centrally learned models.

\end{abstract}


\section{Introduction}
Accurate organ segmentation is critical for many clinical applications~\cite{clinic_1,clinic_2} at different clinical sites, which may have their specific application requirements that concern different organs.
But in the medical images scenario, curating large-scale annotated datasets even within the site is a challenging job, which is expensive and time-consuming. As a result, the scale of the currently labeled datasets is usually limited, thereby limiting the performance of models learned site per se.

\begin{figure}[t]
    \centering
    \includegraphics[width=0.9\textwidth]{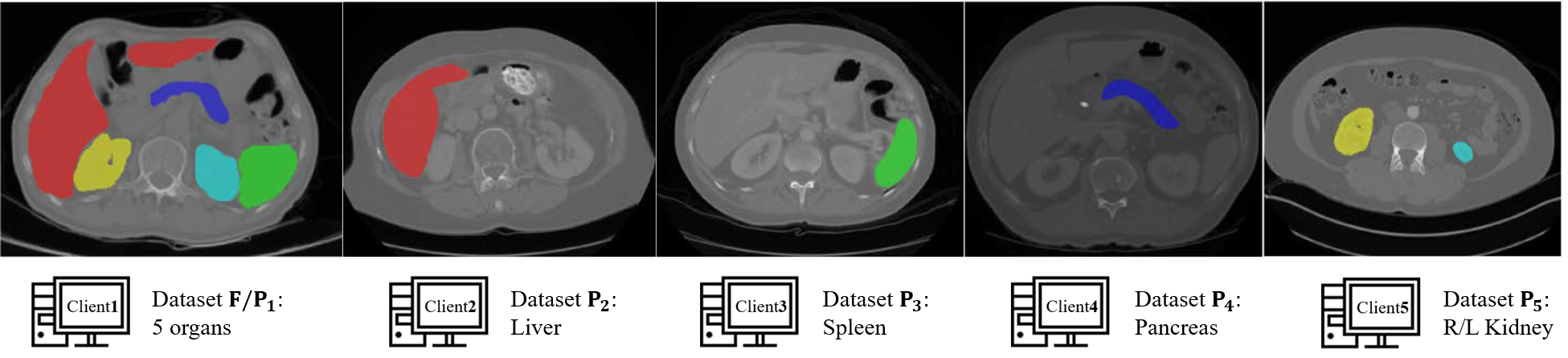}
    \caption{CT images of different partially labeled datasets for different tasks at different clinical sites. The red, green, blue, yellow, and cyan color represent liver, spleen, pancreas, right kidney, and left kidney, respectively.}
    \label{fig1}
\end{figure}

Recently there is a growing interest in learning multi-organ segmentation models from a union of datasets collected from multiple sites, thereby leveraging the strength of `bigger data' for improved performances. Because different sites have site-specific organ labels (as shown in Fig.~\ref{fig1}), forming a partial supervision condition in aggregation.  In other words, most of these datasets are partially annotated in terms of full labeling, which is also a bottleneck in the field of multi-organ segmentation research. 
We can not directly deploy fully supervised segmentation methods on this kind of dataset due to `knowledge conflict'~\cite{liu2022_IL} problem: \textit{The foreground in one dataset becomes background in another dataset}. 
Although some existing studies~\cite{pseg_1,pseg_2,pseg_3,pseg_4,pseg_5,pseg_6,pseg_7} have solved 
this problem, making use of these datasets together, they all conduct model learning centrally.
In reality, 
due to ethical and privacy issues, data cannot be shared among different institutions, nor are data among different departments of the same institution. Data often exists in the form of isolated islands, a phenomenon that is more common in the medical domain. 
Therefore, traditional centralized deep learning methods encounter serious challenges because the data are fragmented into many different sites. 
At the same time, more and more attention has been paid to the security and privacy of the data, especially the security and privacy of medical image data. Many countries have promulgated relevant data security protection laws, such as the EU's General Data Protection Regulation (GDPR)~\cite{yang2019federated}. 
The above issues make the collection and the use of large-scale datasets for machine learning more difficult.

The timely emergency of federated learning (FL) technology alleviates the data privacy issue~\cite{fed_mcmahan}. FL allows that data do not leave their original institution or organization, and train the global federated model collaboratively and effectively under the coordination of a central parameter server. 
During the standard training process of the global federated model, each local client firstly downloads federated model parameters from the central server and trains the model parameters locally based on local data. Next, the locally trained model parameters for each client are sent back to the central server. Then, the central server aggregates model parameters from all clients to get a better global federated model. The final global federated model is obtained through multiple iterations of the above steps. In this process, all data in each client never leave the local storage, but the task in each client \textit{should be same}. 
In ~\cite{federated_new_1}, the authors curate partially labeled datasets by constructing different encoders for different tasks in federated form, which limits the scalability of this method as the clients and tasks increases. Directly encoding multi-task into a single network via FL failed~\cite{federated_new_2} because of `knowledge conflict' problem.
Therefore, the representational capabilities of a single network are far from being exploited. 
In this work, we attempt to bridge the gap by proposing a two-stage approach to multi-site organ segmentation. The first stage of \underline{federated aggregation} is to learn \textit{a unified single multi-organ} segmentation model in a federated fashion 
under \textit{the condition of partial supervision}, with `knowledge conflict' problem solved. 


The goal of multi-site organ segmentation requires learning site-dependent, and application-specific organ segmentation models. Directly using the federated multi-organ segmentation model as the multi-site model is not ideal as it is known that the performance of a federated model is inferior to that of a centrally learned model for the same task. Furthermore, there is a domain shift problem among different sites as they use different scanners and protocols for data acquisition. To accommodate such a domain shift, we propose the second stage of \underline{site adaptation}, which transfers the federated multi-organ segmentation model to a site-dependent, application-specific organ segmentation model in order to further improve the performance of each organ segmentation task per site.

Our main contributions can be summarized as:
\begin{itemize}
    \item We make \textit{the first attempt} in the literature to merge multiple partially labeled datasets of different tasks from different sources into \textit{a unified single model} via federated learning method, fusing different tasks in different clients to one single federated model on the central server and also solving the privacy issue.
    \item We design a new combination of loss functions based on marginal loss and exclusion loss~\cite{pseg_6} to solve the `knowledge conflict' problem and improve the performance of partially supervised multi-organ segmentation.
    \item We transfer the features of the federated model to a local site-dependent, application specific organ segmentation model in order to alleviate the domain gap and further improve the performance of each local model, thereby realizing multi-site organ segmentation. 
   \item  Our extensive experiments verify the effectiveness of our approach on multisite organ segmentation scenario, getting state-of-the-art (SOTA) performance compared with the upper bound method~\cite{pseg_6}, which can centrally access all datasets during training, and surpassing the performance of models trained separated on each local site by a large margin.
\end{itemize}
The rest of this paper is organized as follows: 
In Section~\ref{sec_exps}, extensive experiments are conducted to demonstrate the effectiveness of the method
. Section~\ref{sec_conclusion} 
presents our conclusions. Section~\ref{sec_method} presents the details of the method.

\section{Results}
\label{sec_exps}

We conduct a series of experiments to verify the following assumptions: (i) The different combination of loss functions plays an important role in our Partially Supervised Multi-Organ Segmentation (PSMOS) task, either non-federated learned (\textit{aka} centrally learned, non-FL PSMOS) or federated learned (FL PSMOS). (ii) FL PSMOS can achieve a comparable performance to non-FL PSMOS. (iii) Site adaptation at each local clinical site makes the segmentation performance better.

\subsection{Experiment Settings}

\subsubsection{Datasets and evaluation metrics}
In our experiments, for fair comparison with MargExc~\cite{pseg_6} that is a centrally-learned multi-organ segmentation approach as our baseline model, we use the same five abdominal datasets with different annotation of different organs used in MargExc, including liver, spleen, pancreas, left kidney, and right kidney. 
The details of these datasets, $F$ or $P_1$ (with full label), $P_2$ (with liver label only), $P_3$ (with spleen label only), $P_4$ (with pancrease label only), $P_5$ (with labels of left and right kidneys), are shown in Table~\ref{table_dataset}. 

As the prominent and robust performance of nnUNet~\cite{nnunet} in medical image segmentation tasks, our multi-organ segmentation models are all constructed based on it. We preprocess all datasets to a unified spacing (2.41×1.63×1.63) and normalize them with mean and std of 90.9 and 65.5, respectively. Due to the GPU memory limitation, we crop 3D patches of size [80, 160, 128] from 3D volumes during training. And we only keep the patches whose target area is larger than 33\% for the stability of model training.
There are two patches per iteration and 250 iterations per epoch in our experiments. 
The initial learning rate of the model is 0.01. Whenever the loss reduction is less than 0.001 in consecutive 10 epochs, the learning rate decays by 20\%. All other remaining hyper-parameters, including settings for data augmentation, follow the default settings of the nnUNet~\cite{nnunet}. 

\begin{table*}[t]
    \centering
    \caption{The overview of five benchmark datasets used in our experiments.}\label{t2}
    \resizebox{\linewidth}{!}{
        \begin{tabular}{l|lrrlr}
        \hline
            Datasets & Modality & \# training data & \# testing data & Annotated organs & Mean spacing (z,y,x) in mm\\ \hline
            $F$ ($P_1$)  & CT & 24 & 6 & Fully annotated & (3.00, 0.76, 0.76) \\ 
            $P_2$ & CT & 105 & 26 & Liver & (1.00, 0.77, 0.77) \\
            $P_3$ & CT & 33 & 8 & Spleen & (1.60, 0.79, 0.79) \\
            $P_4$ & CT & 225 & 56 & Pancreas & (2.50, 0.80, 0.80) \\ 
            $P_5$ & CT & 186 & 24 & Left \& right kidneys & (0.80, 0.78, 0.78) \\ \hline
        \end{tabular}
    }
    \label{table_dataset}
\end{table*}

We use the Dice coefficient (DC) and 95$^{th}$ percentile Hausdorff distance (HD95) to evaluate our results.
The higher the DC 
or the smaller the HD95, 
the better the segmentation model.


\subsubsection{Loss functions in training of non-FL PSMOS}
\label{subsec_loss}
In this section, we evaluate the influence of different combinations of loss functions in PSMOS task, and take non-FL format as an example.
The training process starts from training the model for 120 epochs using only the fully-annotated dataset $F$, which provides a good start for the whole training process of multi-organ segmentation model. Then it continues the training for 380 epochs using both the fully-labeled dataset $F$ and the other four partially-labeled datasets, $P_{2,3,4,5}$.
The loss functions used in our experiments include Dice Loss~\cite{dicece}, Lovasz Loss~\cite{lovasz}, Cross Entropy Loss~\cite{dicece}, Top-K Loss~\cite{topk} and Focal Loss~\cite{focal}. According to whether the input CT image is fully annotated or not, we adopt different version of corresponding loss functions. Specifically, for fully-labeled data in $F$, we use conventional loss functions directly; for partially-labeled data in $P_{2,3,4,5}$, we use the marginal and exclusion variants of the Dice Loss, Lovasz Loss, Cross Entropy Loss, Top-K Loss and Focal Loss. The details please refer to Section~\ref{sec_method}.

\subsubsection{Training of FL PSMOS}
We simulate the server ($S$) with one RTX Titan GPU and terminal clients 
($T_n$, $n\in\{1,\ldots,5\}$)with five RTX Titan GPUs. The server owns the pre-trained multi-organ segmentation model, $\Phi_{pretrained}^{m}$, which is obtained by 
training the 3D nnUNet~\cite{nnunet} 
for 120 epochs with the fully-labeled dataset $F$, 
which is owned by the client $T_1$. Other client $T_n$ 
owns partially-annotated dataset $P_{n}$, $n\in\{2,\ldots,5\}$. All clients share the same neural network structure (i.e., 3D nnUNet~\cite{nnunet}) and hyper-parameters.

Finally, we train the following models 
for comparison
.
\begin{itemize}
    \item $\Phi_{Mix}^{m}$: the non-FL multi-organ segmentation model trained 
    centrally in one site;
    \item $\Phi_{Fed}^{m}$: the FL multi-organ segmentation model trained in the form of federated learning;
    \item $\Phi_{F}^{m}$: the multi-organ segmentation model trained only with the fully annotated dataset $F$;
    \item $\Phi_{P_n}^{b}$: the binary segmentation model trained with the partially annotated dataset $P_n$ ($n\in\{2,3,4\}$);
    \item $\Phi_{P_5}^{t}$: The ternary segmentation model trained with the partially labeled dataset $P_5$.
\end{itemize}
The non-FL models  $\Phi_{F}^{m}$, $\Phi_{P_2}^{b}$, $\Phi_{P_3}^{b}$, $\Phi_{P_4}^{b}$ and $\Phi_{P_5}^{t}$ use a combination of conventional Dice Loss~\cite{dicece}, Cross Entropy Loss~\cite{dicece}, and Lovasz Loss~\cite{lovasz}; the non-FL model $\Phi_{Mix}^{m}$ and FL model $\Phi_{Fed}^{m}$ use a mix of conventional and marginal\& exclusion variants of Dice, Cross Entropy, and Lovasz loss functions~\cite{pseg_6}. The details of training the FL PSMOS models, please refer to Section~\ref{sec_method}.

\subsubsection{Adapting the FL model to local sites}
After federated learning, we obtain a FL PSMOS model, $\Phi_{Fed}^{m}$, which is trained utilizing all data from different clients, without private information leakage. Then we adapt $\Phi_{Fed}^{m}$ to each clinical site to suit their own task. 
Since there are three 
ways of fine-tuning for site adaptation, we obtain three models: $\Phi_{FTA}$ (fine-tuning both the encoder and decoder), $\Phi_{FTB}$ (fine-tuning only the decoder), and $\Phi_{FTC}$ (fine-tuning only the encoder).

\subsection{Results}

\begin{table*}[t]
	\centering
	\caption{The performance of the non-FL (aka centrally learned) multi-organ segmentation model 
     $\Phi_{Mix}^m$ on $F$ and $P_*$ datasets, using different combinations of loss functions. $L^{m\&e}_{*}$ means the combination of marginal * loss and exclusion * loss ($L^m_{*}$ + $L^e_{*}$).}
	\newsavebox{\tablebox}
	\resizebox{\linewidth}{!}{	        
        \begin{tabular}{ccccc|rr|rr|rr|rr|rr|rr}
        \hline
        \multicolumn{5}{c}{Different combinations of loss functions} &  \multicolumn{2}{c}{[$F$] Liver} & \multicolumn{2}{c}{[$F$] Spleen} & \multicolumn{2}{c}{[$F$] Pancreas} & \multicolumn{2}{c}{[$F$] L Kidney} &\multicolumn{2}{c}{[$F$] R Kidney} & \multicolumn{2}{c}{Mean}\\
				\hline
			$L^{m\&e}_{Dice}$&$L^{m\&e}_{Lovasz}$&$L^{m\&e}_{CE}$&$L^{m\&e}_{Topk}$&$L^{m\&e}_{Focal}$& DC & HD95 & DC & HD95 & DC & HD95 & DC & HD95 & DC & HD95 & DC & HD95\\
        \hline
        \Checkmark&&\Checkmark&& & .969&1.71  & .943&1.47   & .835&3.27     & .945&1.58    & .950&1.35    & .928&1.88 \\ 
        \Checkmark&&&&\Checkmark  & \textbf{.972}&1.77  & .935&1.62   & .826&4.43     & .944&1.40    & .949&1.22    & .925&2.09 \\ 
        \Checkmark&&&\Checkmark&  & .971&\textbf{1.59} & .931&1.60   & .826&4.67     & .948&1.37    & \textbf{.953}&\textbf{1.21}    & .926&2.09 \\ 
        &\Checkmark&\Checkmark&&  & .969&1.65  & .940&1.50   & .834&3.92     & .943&1.39    & .949&1.24    & .927&1.94 \\ 
        \hline
        \Checkmark&&\Checkmark&&\Checkmark   & .970&1.91  & .945&1.37   & .833&3.79     & .943&\textbf{1.34}    & .951&1.95    & .928&3.71 \\ 
        \Checkmark&&\Checkmark&\Checkmark&  & \textbf{.972}&\textbf{1.59}  & \textbf{.947}&\textbf{1.36}   & .824&6.71     & .948&1.37    & \textbf{.953}&1.28    & .929&2.46 \\ 
        \Checkmark&\Checkmark&\Checkmark&&  & \textbf{.972}&1.79  & .946&1.43   & \textbf{.849}&\textbf{3.04}     & \textbf{.950}&1.37    & .948&1.28    & \textbf{.933}&\textbf{1.78} \\ \hline
               \hline
        \multicolumn{5}{c}{Different combinations of loss functions} &  \multicolumn{2}{c}{[$P_2$] Liver} & \multicolumn{2}{c}{[$P_3$] Spleen} & \multicolumn{2}{c}{[$P_4$] Pancreas} & \multicolumn{2}{c}{[$P_5$] L Kidney} &\multicolumn{2}{c}{[$P_5$] R Kidney} & \multicolumn{2}{c}{Mean}\\
				\hline
	$L^{m\&e}_{Dice}$&$L^{m\&e}_{Lovasz}$&$L^{m\&e}_{CE}$&$L^{m\&e}_{Topk}$&$L^{m\&e}_{Focal}$& DC & HD95 & DC & HD95 & DC & HD95 & DC & HD95 & DC & HD95 & DC & HD95\\
        \hline
        \Checkmark&&\Checkmark&&  & .941&9.57  & \textbf{.975}&\textbf{1.05}   & .790&6.08     & .972&1.64    & .966&15.61    & .929&6.79 \\ 
        \Checkmark&&&&\Checkmark  & .935&13.30  & .970&1.07   & .778&6.13     & .971&4.26    & .957&16.02    & .922&8.16 \\ 
        \Checkmark&&&\Checkmark&  & .928&15.23  & .972&1.10   & .779&6.15     & .954&2.89    & .949&15.48    & .916&8.17 \\ 
        &\Checkmark&\Checkmark&&  & .940&13.11  & .969&1.08   & .788&6.29     & .972&3.56    & .954&15.38    & .925&7.88 \\ 
        \hline
        \Checkmark&&\Checkmark&&\Checkmark  & .942&13.12  & \textbf{.975}&\textbf{1.05}   & .786&6.05     & .970&3.20    & .968&15.40    & .928&8.74 \\ 
        \Checkmark&&\Checkmark&\Checkmark&  & .937&9.78  & .972&\textbf{1.05}   & .783&6.26     & .970&5.06    & .950&17.41    & .922&8.95 \\ 
        \Checkmark&\Checkmark&\Checkmark&&  & \textbf{.948}&\textbf{6.54}  & .974&\textbf{1.05}   & \textbf{.795}&\textbf{5.08}     & \textbf{.974}&\textbf{1.44}    & \textbf{.969}&\textbf{14.57}    & \textbf{.932}&\textbf{6.49} \\ \hline
        \end{tabular}
	}
\label{table_lossresult}
\end{table*}

\begin{table*}[t]
	\centering
	\caption{Comparison with state-of-the-art results. The Training/Testing set split of ours is the same as compared methods. The non-FL, FL and FT means centrally trained, federated trained, and adapted or fine-tuned model, respectively.}
        \resizebox{\linewidth}{!}{
    	\begin{tabular}{l|rr|rr|cc|cc|cc|cc}
            \hline
            \multirow{2}*{\centering \diagbox{Methods}{Organs}} &  \multicolumn{2}{c}{[$F$\&$P_2$] Liver} & \multicolumn{2}{c}{[$F$\&$P_3$] Spleen} & \multicolumn{2}{c}{[$F$\&$P_4$] Pancreas} & \multicolumn{2}{c}{[$F$\&$P_5$] L Kidney} &\multicolumn{2}{c}{[$F$\&$P_5$] R Kidney} & \multicolumn{2}{c}{Mean}\\
    				\cline{2-13}
    			& DC & HD95 & DC & HD95 & DC & HD95 & DC & HD95 & DC & HD95 & DC & HD95\\
            \hline
            PaNN~\cite{pseg_4}  & \textbf{.961}&\textbf{2.99}  & .941&11.21   & .767&7.04     & .919&3.76    & .943&1.43    & .906&5.29 \\ 
            PIPO~\cite{pseg_5}  & .940&10.13  & .919&11.54   & .772&6.08     & .948&4.56    & \textbf{.962}&\textbf{1.17}    & .908&6.69 \\ 
            MargExc~\cite{pseg_6}  & .955&5.64  & .959&1.26   & .813&4.68     & .959&1.61   & .958&8.48    & .929&4.33 \\ 
            \hline
            $\Phi_{Mix}^{m}$ (Ours, non-FL)  & .960&4.17  & \textbf{.960}&\textbf{1.24}   & \textbf{.822}&\textbf{4.06}     & \textbf{.962}&\textbf{1.41}    & .959&7.93    & \textbf{.933}&\textbf{3.76} \\ 
            $\Phi_{Fed}^{m}$ (Ours, FL)  & .952 & 4.76  & .952&1.77   & .814&5.89     & .952&2.56    & .949&9.03    & .924 & 4.80\\
            $\Phi_{FTB}$ (Ours, FT)  & .957 & 4.00  & .958 & 1.94   & .816&5.24     & .959 & 2.34    & .953&8.36    & .929 & 4.38\\
            \hline
            \end{tabular}
            }
	\label{table_noflvssota}
\end{table*}

\begin{figure*}[t]
    \centering
    \includegraphics[width=\textwidth]{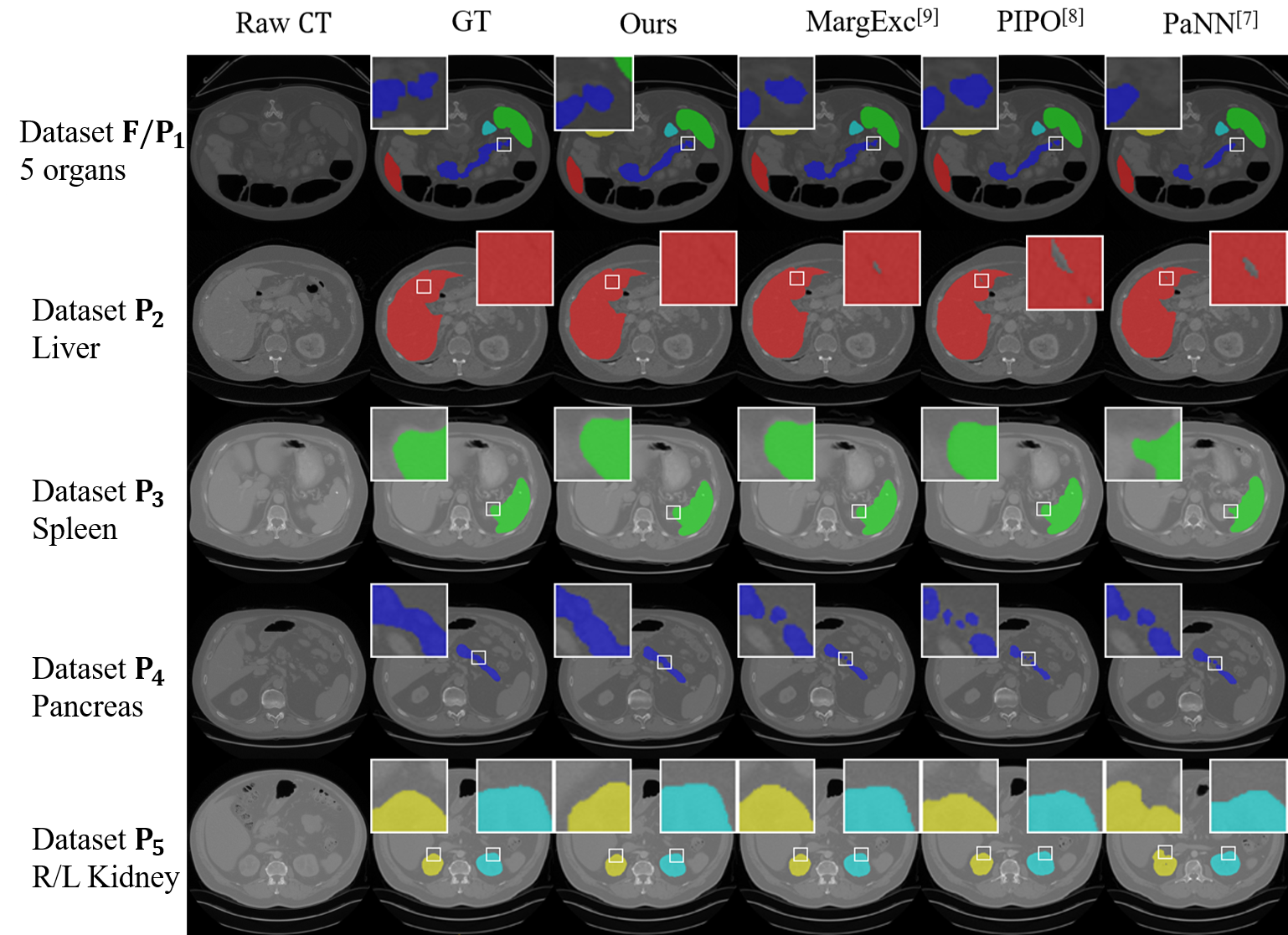}
    \caption{Visualization of the segmentation results of our proposed method and other SOTA methods.}
    \label{fig:non-flvssota}
\end{figure*}

\begin{table*}[t]
	\centering
	\caption{The segmentation performance comparison between models trained separately ($\Phi_{F}^{m}$, $\Phi_{P_n}^{*}$), centralized ($\Phi_{Mix}^{m}$) and federated ($\Phi_{Fed}^{m}$).}
\resizebox{\linewidth}{!}{
		\begin{tabular}{l|rr|rr|rr|rr|rr|rr}
        \hline
        \multirow{2}*{\centering \diagbox{Methods}{Organs}} &  \multicolumn{2}{c}{[$F$] Liver} & \multicolumn{2}{c}{[$F$] Spleen} & \multicolumn{2}{c}{[$F$] Pancreas} & \multicolumn{2}{c}{[$F$] L Kidney} &\multicolumn{2}{c}{[$F$] R Kidney} & \multicolumn{2}{c}{Mean}\\
				\cline{2-13}
			& DC & HD95 & DC & HD95 & DC & HD95 & DC & HD95 & DC & HD95 & DC & HD95\\
        \hline
        $\Phi_{F}^{m}$   & .950&2.94  & .925&4.35   & .817&6.02     & .918&2.47    & .915&2.79    & .905&3.71 \\ 
        \hline
        $\Phi_{Fed}^{m}$ (Ours) & .965&2.41  & .939&2.33   & .840&4.79     & .938&1.65    & .936&1.50    & .924&2.54 \\ 
        $\Phi_{Mix}^{m}$ (Ours, Upper bound)  & .972&1.79  & .946&1.43   & .849&3.04     & .950&1.37    & .948&1.28    & .933&1.78 \\ \hline \hline
            \multirow{2}*{\centering \diagbox{Methods}{Organs}} &  \multicolumn{2}{c}{[$P_2$] Liver} & \multicolumn{2}{c}{[$P_3$] Spleen} & \multicolumn{2}{c}{[$P_4$] Pancreas} & \multicolumn{2}{c}{[$P_5$] L Kidney} &\multicolumn{2}{c}{[$P_5$] R Kidney} & \multicolumn{2}{c}{Mean}\\
				\cline{2-13}
			& DC & HD95 & DC & HD95 & DC & HD95 & DC & HD95 & DC & HD95 & DC & HD95\\
			\hline
            $\Phi_{P_2}^{b}$  & .914&7.02 & -     &-& -     &-& -  &  -     & -    &  -   & \multirow{4}{*}{.894} &\multirow{4}{*}{8.02}\\ 
            $\Phi_{P_3}^{b}$  & -    &- & .955& 1.41& -     &-& -   &   -   & -    &   -  &   &                     \\ 
            $\Phi_{P_4}^{b}$  & -     &-& -     &-& .747 &7.63& - &   -     & -    &   -  &  &                      \\ 
            $\Phi_{P_5}^{t}$  & -     &-& -     &-& -     &-& .928 &  5.57  & .926  &  18.50 &  &                      \\ 
            \hline
            $\Phi_{Fed}^{m}$ (Ours) & .938&7.10 & .965&1.21 & .788&6.99 & .966&   3.47  & .961   & 16.56 & .924  &7.06                \\ 
            $\Phi_{Mix}^{m}$ (Ours, Upper bound)  & .948&6.54 & .974&1.05 & .795& 5.08 & .974&  1.44   & .969    &14.57 & .932  &5.74                \\ 
            \hline
                    \end{tabular}
         }
	\label{table_flf}
\end{table*}


\begin{table}[t]
    \centering
    \caption{Experimental results under different federated learning hyper-parameters settings.}\label{table_flablation}
    \begin{lrbox}{\tablebox}
        \begin{tabular}{lrr|r|rrr}
        \hline
        \multicolumn{4}{c}{Different training setting}&\multicolumn{3}{c}{Results}\\
        \cline{1-7}
                  Model & Client iteration & Global round (epoch) & Total iteration & Dice  & HD95  & Training time (h)\\ 
                   \hline
        $\Phi_{Fed}$ V1  & 25            & 4,000 &   \multirow{3}{*}{100,000}      & .895 & 5.44 & 94      \\ 
        $\Phi_{Fed}$ V2  & 50            & 2,000  &         & .924 & 4.80 & 89      \\ 
        $\Phi_{Fed}$ V3  & 100           & 1,000   &   & .919 & 4.61   & 82      \\ \hline
        $\Phi_{Mix}$(non-FL)   & 250           & 500 & 125,000           & .933 & 3.76  & 75      \\ \hline
        \end{tabular}
    \end{lrbox}
    \scalebox{0.85}[0.85]{\usebox{\tablebox}}
\end{table}

\subsubsection{The results of non-FL PSMOS model}\label{subsec_loss_res}
The performance of non-FL PSMOS models with different combinations of loss functions on $F$ and $P_{2,3,4,5}$ datasets is shown in Table~\ref{table_lossresult}. We observe that the 
centrally trained $\Phi_{Mix}^m$ achieves the best performance 
with the combination of $\{L^m_{Dice}$, $L^m_{CE}$, $L^m_{Lovasz}$, $L^e_{Dice}$, $L^e_{CE}$, $L^e_{Lovasz}\}$. The average DC 
and HD95 reach 0.933 and 1.78, respectively, on the fully annotated dataset $F$, 
and 0.932 and 6.49, respectively, on the partially-annotated dataset $P_{n},(n\in\{2,3,4,5\})$. In the first four rows of two tables in Table~\ref{table_lossresult}, which are the combinations of two loss functions. Based on the basic combination of Dice and CE loss in MargExc~\cite{pseg_6}, we try different combinations and find that Dice and CE loss are indeed the best combination. Based on this discovery, we add other loss functions separately and find that $\{L^{m\&e}_{Dice}$, $L^{m\&e}_{CE}$, $L^{m\&e}_{Lovasz}\}$ is the best combination scheme in the end, which indicate that improve the combination of loss functions plays an important role in the PSMOS task. In particular, the optimized loss function have a significant improvement on small and irregular organs such as the pancreas.


Compared with other SOTA methods as shown in Table~\ref{table_noflvssota}, our model $\Phi_{Mix}^m$ achieves better performance. Specifically, the average DC of the proposed method using the above-mentioned loss function combination is 2.7\%, 2.5\% and 0.4\% higher than that of PaNN~\cite{pseg_4}, PIPO~\cite{pseg_5}, and MargExc~\cite{pseg_6}, respectively. At the same time, our method also greatly reduces the HD95, which is 28.9\%, 43.8\%, and 13.2\% less than the other three methods, respectively. From the visualization results shown in Fig.~\ref{fig:non-flvssota}, we can also perceive that our method works better, especially on small and irregular organ such as pancreas, which have finer segmentation results compared with other SOTA methods.

\subsubsection{The results of FL PSMOS model}
In FL setting, we treat non-FL PSMOS model ($\Phi_{Mix}^m$) as upper bound for comparison. Because $\Phi_{Mix}^m$ is trained centrally in one site, which is the most ideal scenario for training. In Table~\ref{table_flf}, the average DC of $\Phi_{Mix}^m$ on $F$ and $P_n, n\in\{2,3,4,5\}$ are 0.933 and 0.932, respectively, and HD95 is 1.78 and 5.74, respectively.

When we do not aggregate these partially labeled datasets together, there is no data leakage problem either, but the performance of the segmentation models suffer greatly due to the greatly reduced scale of the dataset. For instance, in $T_1$, $\Phi_{F}^m$ is trained based on its own dataset $F$ and has no access to other dataset in other clients. The performance of $\Phi_{F}^m$ on its own dataset $F$ reachs 0.905 and 3.71 for DC and HD95, respectively. Similarly, $\Phi_{P_n}^b$, $n\in\{2,3,4\}$ and $\Phi_{P_5}^t$ are also trained on their own partially labeled datasets $P_n$, $n\in\{2,3,4,5\}$, achieving average DC and HD95 of 0.894 and 8.02, respectively. The performances of these models trained separately on their own dataset all have a large gap compared with $\Phi_{Mix}^m$.

When we involve FL, $\Phi_{Fed}^m$ can be trained on all datasets from all clients without data leakage, aggregating knowledge from all data that can be accessed. And in our setting, the tasks of different clients are different. We also aggregate these tasks into a single multi-organ segmentation model, and the `knowledge conflict' problem between different tasks in partially annotated datasets is also solved through the combination of loss functions discussed in Section~\ref{subsec_loss_res}. In Table~\ref{table_flf}, $\Phi_{Fed}^m$ achieves 0.924, 2.54 (and 0.924, 7.06) for average DC and HD95 on $F$ (and $P_n$), respectively. 
MargExc~\cite{pseg_6} concludes that a single model trained on a single source can not generalized well to other sources based on the results in Table 3~\cite{pseg_6}. 
The $\Phi_{Fed}^m$ can surpass $\Phi_{F}^m$ and $\Phi_{P_n}$ by a large margin, indicating the strength of `bigger data' from multi-site. 
That $\Phi_{Fed}^m$ has a comparable performance with $\Phi_{Mix}^m$ also demonstrate the effectiveness of our multi-site approach. In Table~\ref{table_noflvssota}, our $\Phi_{Fed}^m$ outperforms other centrally trained models~\cite{pseg_4,pseg_5}.

In FL, we can set how many iterations to train on each client before combining parameters to the server. In Table~\ref{table_flablation}, as expected, the fewer iterations on each client before aggregation, the more time-consuming of the overall training. As the Global epoch decreases, the performance of the segmentation model first increases and then decreases. In our experiments, under the same total training iterations, we choose 50 and 2,000 for Client iteration and Global epoch, respectively.


\subsubsection{The results of site-adapted model}

As mentioned above, we introduce 
three ways of fine-tuning for site adaptation in each client, thus obtaining three models: (i) $\Phi_{FTA}$, which fine-tunes both the encoder and decoder of the segmentation model; (ii) $\Phi_{FTB}$ fine-tunes only the decoder; (iii) $\Phi_{FTC}$ fine-tunes only the encoder.

As shown in Table~\ref{table_ftf}, we observe that both $\Phi_{FTA}$ and $\Phi_{FTB}$ contribute to the performance of the original FL model $\Phi_{Fed}^{m}$, especially the model $\Phi_{FTB}$. 
On the dataset $F$ (and $P_n$), for average DC and HD95, the $\Phi_{FTB}$ scheme improves the performance of $\Phi_{Fed}^m$ to 0.929, 2.31 (and 0.928, 6.44), respectively. 
In Table~\ref{table_noflvssota}, our $\Phi_{FTB}$ further approaches to the best non-FL model $\Phi_{Mix}^m$.


\begin{table*}[t]
	\centering
	\caption{The segmentation performances of the site-adapted models.}
	   \resizebox{\linewidth}{!}{
		\begin{tabular}{l|l|rr|rr|rr|rr|rr|rr}
        \hline

        \multirow{2}{*}{Model}&\multirow{2}*{\centering \diagbox{train set}{test set}} &  \multicolumn{2}{c}{[$F$] Liver} & \multicolumn{2}{c}{[$F$] Spleen} & \multicolumn{2}{c}{[$F$] Pancreas} & \multicolumn{2}{c}{[$F$] L Kidney} &\multicolumn{2}{c}{[$F$] R Kidney} & \multicolumn{2}{c}{Mean}\\
				\cline{3-14}
			&  & DC & HD95 & DC & HD95 & DC & HD95 & DC & HD95 & DC & HD95 & DC & HD95\\
        \hline
        $\Phi_{Fed}^{m}$   &[$F\&P_{2:5}$]    & .965&2.41  & .939&2.33   & \textbf{.840}&4.79     & .938&1.65    & .936&1.50    & .924&2.54 \\ \hline
        $\Phi_{FTA}$   &\multirow{3}{*}{[$F$]}    & .967&\textbf{1.78}  & .933&3.25   & \textbf{.840}&4.03     & .945&1.58    & \textbf{.944}&\textbf{1.43}    & .925&2.41 \\ 
        $\Phi_{FTB}$   &    & \textbf{.969}&\textbf{1.78}  & \textbf{.946}&\textbf{2.74}   & .838&\textbf{3.97}     & \textbf{.946}&\textbf{1.63}    & \textbf{.944}&\textbf{1.43}    & \textbf{.929}&\textbf{2.31} \\ 
        $\Phi_{FTC}$   &    & .960&2.22  & .925&5.60   & .831&4.59     & .937&2.06    & .938&1.87    & .918&3.27 \\ \hline \hline
        \multirow{2}{*}{Model}&\multirow{2}*{\centering \diagbox{train set}{test set}} &  \multicolumn{2}{c}{[$P_2$] Liver} & \multicolumn{2}{c}{[$P_3$] Spleen} & \multicolumn{2}{c}{[$P_4$] Pancreas} & \multicolumn{2}{c}{[$P_5$] L Kidney} &\multicolumn{2}{c}{[$P_5$] R Kidney} & \multicolumn{2}{c}{Mean}\\
				\cline{3-14}
			&  & DC & HD95 & DC & HD95 & DC & HD95 & DC & HD95 & DC & HD95 & DC & HD95 \\
        \hline
        $\Phi_{Fed}$   &[$F\&P_{2:5}$]    & .938&7.10  & .965&1.21  & .788&6.99     & .966&3.47    & .961&16.56  & .924 & 7.06   \\ \hline
        \multirow{9}{*}{$\Phi_{FTA}$}   &\multirow{3}{*}{[$P_2$]}    & \textbf{.945}&6.55  & -&-   & -&-     & -&-    & -&- & \multirow{9}{*}{.927}&\multirow{9}{*}{6.62} \\ 
        \multirow{9}{*}{$\Phi_{FTB}$}   &    & \textbf{.945}&\textbf{6.22}  & -&-   & -&-     & -&-    & -&-  & \multirow{9}{*}{\textbf{.928}}&\multirow{9}{*}{\textbf{6.44}}\\ 
        \multirow{9}{*}{$\Phi_{FTC}$}   &     & .938&6.83  & -&-   & -&-     & -&-    & -&-& \multirow{9}{*}{.915}&\multirow{9}{*}{6.91} \\  
        \cline{2-12}
           &\multirow{3}{*}{[$P_3$]}     & -&-  & .965&1.20   & -&-     & -&-    & -&- & &   \\  
           &     & -&-  & \textbf{.970}&\textbf{1.14}   & -&-     & -&-    & -&- & & \\
           &     & -&-  & .959&1.20   & -&-     & -&-    & -&-& &  \\ 
            \cline{2-12}
           &\multirow{3}{*}{[$P_4$]}  & -&-     & -&-   & .780&6.71      & -&-    & -&-  & &  \\
           &  & -&-     & -&-   & \textbf{.793}&\textbf{6.50}      & -&-    & -&-  & &\\
           &  & -&-     & -&-   & .760&6.88      & -&-    & -&-   & &\\ 
        \cline{2-12}
           &\multirow{3}{*}{[$P_5$]}   & -&-   & -&-     & -&-   & \textbf{.974}&\textbf{2.59} & \textbf{.969}&16.03& &   \\ 
           &    & -&-   & -&-     & -&-   & .972&3.04  & .962&\textbf{15.29} & &  \\
          &    & -&-   & -&-     & -&-    & .963&3.11  & .957&16.54 & & \\ \hline
                \end{tabular}}
	\label{table_ftf}
\end{table*}

\section{Discussion}
\label{sec_conclusion}

Clinical practice requires that the learning of organ segmentation models should protect data privacy and 
can not rely too much on a large amount of fully labeled images, because these annotations are not easy to obtain. 
In this work, we propose to learn a multi-site multi-organ segmentation model trained on partially labeled datasets from different clinical sites with different target tasks via federated partial supervision. This model aggregates knowledge from all datasets at different sites into a single multi-organ segmentation model, but without any data leakage from each clinical site. Based on this federated learned model with powerful representation capabilities, each clinical site can further adapt this ability to its own needs and applications. 
We also find the optimal combination of loss functions to maximize knowledge utilization in the PSMOS task, and the model trained by this combination is also considered as our upper bound performance, $\Phi_{Mix}^m$. Extensive experiments have verified the effectiveness of our Federated Partially Supervision method, which can reach the performance of upper bound. In each site, the fine-tuned model can further get a better performance. Our solution can effectively relieve the labeling burden for doctors. 

Although our FL model is proven effective compared to the existing methods, it still has several limitations. 
Our model is trained based on partially labeled datasets belonging to the same region of our body, \textit{i.e.}, abdominal organs. If we want to train a single FL model including more categories of organs from different regions of our body~\cite{liu2022universal}, we may encounter the problem that the data distribution varies too much between different sites. 
This is also a direction we plan to try in the future. 
Another limitation is that our model training is synchronized between different sites. Every time a new site is added, we must be able to access the data of all past sites at the same time to train a new federated model. There is no data leakage issue, but it is also difficult to access old data from all sites at the same time. Asynchronizing our federated learning is also one of our future research directions, such as through incremental learning~\cite{liu2022_IL}.

\section{Materials and Methods}
\label{sec_method}
\begin{figure}[t]
    \centering
    \includegraphics[width=0.65\columnwidth]{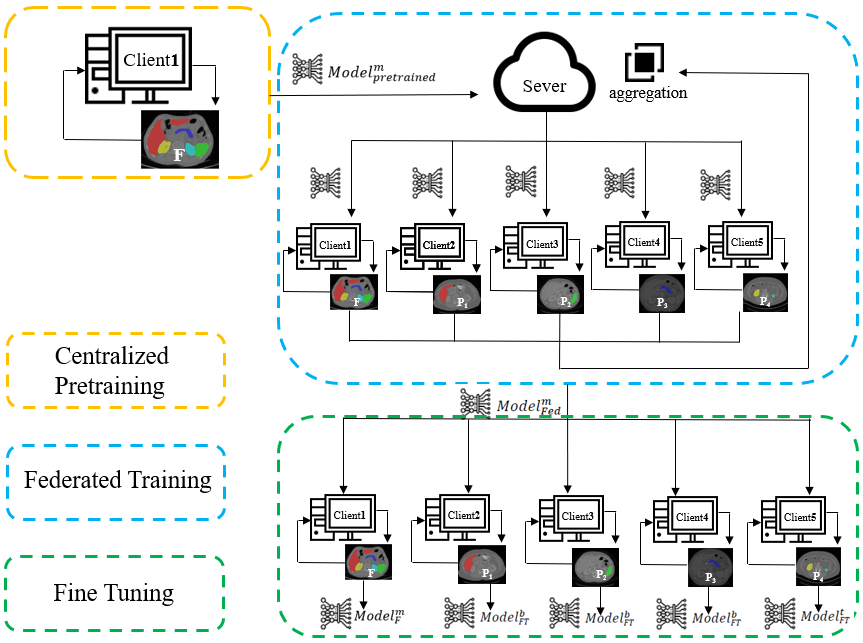}
    \caption{The pipeline of our framework.}
    \label{fig_pipeline}
\end{figure}

Here we describe the technical details of the FL for the partially supervised multi-organ segmentation and site adaptation approach utilized in this work. The proposed method includes two-stage: (i) Federated aggregation that learns a multi-organ segmentation model from partial supervision, and (ii) Site adaptation that fine-tunes the model
with local data to a site-dependent model. Both FL and fine-tuning procedures presented are designed for 3D medical image segmentation tasks. The whole pipeline of our framework is shown in Fig.~\ref{fig_pipeline}

\IncMargin{1em} 
\begin{algorithm}[t]

    \SetAlgoNoLine 
    \SetKwInOut{Input}{\textbf{Input}}
    \SetKwInOut{Output}{\textbf{Output}} 

    \Input{number of clients $K$, FL global rounds $R$, amount of total training iterations $N$ (n iterations each round), learning rate $\eta$\\
    initialize the server model weights: $w^{0}$}
    \Output{optimal $w^{r}$}
    \BlankLine

    \textbf{FL Procedure} \textit{ServerUpdate}:\\
    \For {$r\leftarrow 1$ \KwTo $R$}{ 
    	    \For{client $k\leftarrow 1$ \KwTo $K$}{
                $w_{k}^{r-1} \leftarrow w^{r-1}$\;
    	        $w_{k}^{r}\leftarrow ClientUpdate(k, w_k^{r-1})$
    	    }
         $w^r\leftarrow \sum_{k=1}^{K}\frac{n_{k}}{N}w_{k}^{r}$
     	 }
       \BlankLine
    \textbf{Function} \textit{ClientUpdate} $(k, w_k^{r-1})$:\\
    \For{iteration $i\leftarrow 1\ \KwTo\ n$}{ 
    	    $w_k^{r-1,i}\leftarrow w_k^{r-1,i-1}-\eta \nabla l(w_k^{r-1,i-1})$\;
     	 }
      \textbf{return} $w_k^{r} \leftarrow w_k^{r-1,n}$
    \caption{FL for partial supervised multi-organ segmentation}
     \label{algo_fl} 
\end{algorithm}
\DecMargin{1em}

\subsection{Partially supervised multi-organ segmentation (PSMOS)}
According to the idea of marginal loss and exclusion loss functions in the literature~\cite{pseg_6}, the marginal loss merges all unlabeled organ pixels into the background label, and the exclusion loss uses the pixels of different organs to be mutually exclusive, that is, a pixel can only belong to a certain organ, but cannot belong to multiple organs. We use a variety of common segmentation loss functions, including Dice Loss~\cite{dicece}, Cross Entropy Loss~\cite{dicece}, Focal Loss~\cite{focal}, Top-K Loss~\cite{topk}, Lovasz Loss~\cite{lovasz}, and the combined form of these loss functions, utilizing a small amount of fully annotated data and a large amount of partially annotated data, to train a high-performance multi-organ segmentation neural network in a federated fashion.

For a multi-organ segmentation task with $N$ labels, we define its label set as $\Omega_N=\{C_1,C_2,...,C_N\}$. Suppose that a pixel in segmentation belongs a certain class $C_n$, its label is encoded as a $N$-dimensional one-hot vector $y_{n}^{'} = [y_1,y_2,...,y_N]$, where $y_n=1$, otherwise $0$. Multi-class classifiers contain a set of response functions \{$f_n(x),n\in\Omega_N$\}, and the posterior classification probability is calculated by the softmax function,

\begin{equation}
    p_n = \frac{exp(f_n)}{\sum_{k=1}^{N}exp(f_k)},n\in\Omega_N.
\end{equation}

\subsubsection{Marginal loss}
Assume that for images with incomplete segmentation labels, there are only $M (M<N)$ labels, its corresponding label index set is $\Omega_{M}^{'}$. For each merged class label $C_m\in\Omega_{M}^{'}$, there is a corresponding subset $\Phi_m\subset\Omega_N$. The probability for the merged class $m$ is a marginal probability:
\begin{equation}
    q_m = \sum_{n\in \Phi_m} p_n.
\end{equation}
Also, the one-hot vector for a class $m$ is encoded as $y_{m}^{'} = [y_1,y_2,...,y_M]$. 
Using the above marginal probability $q_m$, we define the corresponding marginal loss for Dice loss, cross entropy loss, focal loss, Top-$K$ loss, and Lovasz loss:
\begin{align}
        L^m_{Dice}=& \sum_{m\in \Omega_{M}^{'}}(1-2\cdot \frac{y_m~q_m}{y_m+q_m})\label{l_m_dice},\\
	    L^m_{CE}=&-\sum_{m\in\Omega_{M}^{'}}y_m\log_{}{q_m}\label{l_m_ce},\\
	    L^m_{Focal}=&-\sum_{m \in \Omega_{M}^{\prime}} y_{m}\left(1-q_{m}\right)^{\gamma} \log \left(q_{m}\right)\label{l_m_focal},\\
	    L^m_{Topk}\left(l_{Z}(f)\right)=&\frac{1}{k} \sum_{i=1}^{k} L^m_{CE[i]}(f)\label{l_m_topk},\\
	    L^m_{Lovasz}=&\frac{1}{m} \sum_{m \in \Omega_{M}^{\prime}} J(e(m))\label{l_m_lovasz}.
\end{align}
In Eq.~(\ref{l_m_focal}), $\gamma$ is parameter of focusing, $(1-q_m)^\gamma$ is a modulation coefficient (we set $\gamma=2$ in our experiment). In Eq.~(\ref{l_m_topk}), $L_{z}(f)=\left\{l_{i}(f)\right\}_{i=1}^{n}$, $l_{i}(f)=l\left(f\left(x_{i}\right), y_{i}\right)$, $l_{[k]}(f)$ is defined as the $k$th largest element in $l_{[i]}(f)$, we set $k=10$ in our experiment. In Eq.~(\ref{l_m_lovasz}), $J$ is the Lovasz extension of Intersection-over-Union (IoU), and $e(m)$ is the misclassification vector for class $m$.

\subsubsection{Exclusion loss}
We define the exclusive subset of class $n$ as $E_n$, and its label is encoded as a non-one-hot $N$-dimensional vector $e_{n}^{'}=[e_1,e_2,...,e_N]$, expressed as follows:
\begin{equation}
    e_{n}^{'}=\sum_{k \in E_{n}} y_{k}^{'}.
\end{equation}

The intersection of the prediction $p_n$ of the segmentation network and exclusive label $e_n$ should be as small as possible. Using the above vector $e^{'}_n$, we define the corresponding exclusion loss for Dice loss, cross entropy loss, focal loss, Top-$K$ loss, and Lovasz loss:  
\begin{align}
        L^e_{Dice}=& \sum_{n \in \Omega_{N}} 2 \cdot \frac{e_{n} p_{n}}{e_{n}+p_{n}}\label{l_e_dice},\\
	    L^e_{CE}=&\sum_{n \in \Omega_{N}} e_{n} \log \left(p_{n}+\varepsilon\right)\label{l_e_ce},\\
	    L^e_{Focal}=&\sum_{n \in \Omega_{N}} e_{n}\left(1-p_{n}\right)^{\gamma} \log \left(p_{n}+\varepsilon\right)\label{l_e_focal},\\
	    L^e_{Topk}\left(l_{Z}(f)\right)=&-\frac{1}{k} \sum_{i=1}^{k} L^e_{CE[i]}(f)\label{l_e_topk},\\ 
	    L^e_{Lovasz}=&-\frac{1}{n} \sum_{n \in \Omega_{N}} J(e(n))\label{l_e_lovasz}.
\end{align}

In Eqs.~(\ref{l_e_ce}) and (\ref{l_e_focal}), we set $\varepsilon=1$ to avoid the trap of $-\infty$.


\subsection{Federated learning for PSMOS}
In our work, the federated learning framework follows the conventional settings as ~\cite{fed_mcmahan}, called client-server setup. In the setting, the coordinator is the server that can send the initial model to each client. Then each client trains the same model architecture locally on their data. Once a certain number of clients finish a round of local training, they send the updated model weights or their gradients to the server. The server collects model weights or gradients from clients simultaneously, then aggregates those gradients. After aggregation, the server re-distributes the new weights to the clients, and the next round of local model training begins. This is called one federated training round, the process is repeated until the model converges or the maximum number of iterations is reached. Each client is allowed to select its local best model by monitoring a certain performance metric on a local validation set. The client can select either the global model returning from the server after averaging or any intermediate model considered best during local training based on their validation metric. The definition of FL is as follows, assuming that there are $K$ clients participating in FL, $n_k$ represents the sample numbers of the client $k$, $n=\sum_{k=1}^{K}n_{k}$ represents the total sample numbers of all clients. The goal of FL is to optimize the following objective loss function:
\begin{equation}
    min~F(w):=\sum_{k=1}^{K}\frac{n_{k}}{n}F_{k}(w),
\end{equation}
where $w$ is the parameter that the model needs to learn, and $F_k$ is the local target loss function of client $k$.
In our experiments, we implement the FederatedAveraging algorithm proposed in~\cite{fed_mcmahan}. The entire FL algorithm for partially supervised multi-organ segmentation is shown in Alg.~\ref{algo_fl}.

In this work, we use a combination of marginal Dice loss, marginal cross entropy loss, marginal Lovasz Loss, exclusion Dice loss, exclusion cross entropy loss, and exclusion Lovasz Loss for a PSMOS task.

\begin{figure}[t]
    \centering
    \includegraphics[width=0.45\columnwidth]{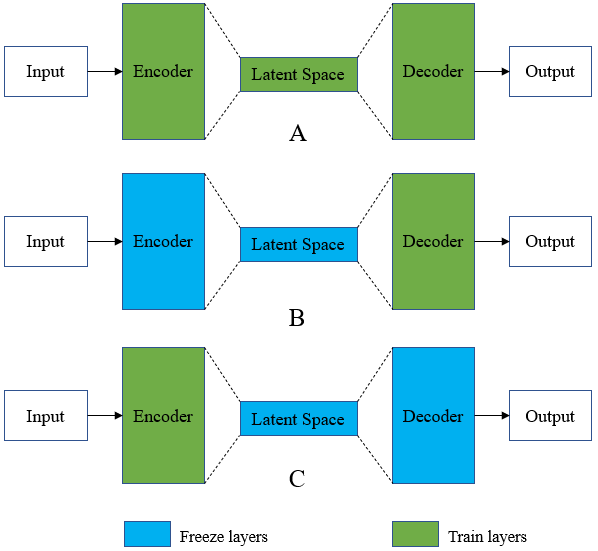}
    \caption{Three different methods of fine-tuning the FL model.}
    \label{fig:finetune}
\end{figure}

\subsection{Site adaptation of the FL model}
Suppose that the federated model is denoted as $\Phi_{Fed}$, which captures the common task knowledge learned from the aggregated data, the site adaptation module transfers this knowledge from mixed to target site for further improving the performance on the target data domain. Such an adaptation is from a \textit{site-independent} multi-organ segmentation model to a \textit{site-dependent, application-specific} organ segmentation model.

As shown in Fig.~\ref{fig:finetune}, we have designed three fine-tuning methods for site adaptation: (i) No freezing. We initialize the target network with the $\Phi_{Fed}$ model parameters and fine-tune the entire neural network. (ii) Freezing encoder. We initialize the target network with the parameters of the $\Phi_{Fed}$ model, freeze the encoder, and fine-tune the decoder part of the network. (iii) Freezing decoder. We initialize with the parameters of the $\Phi_{Fed}$ model, freeze the decoder, and fine-tune the encoder part of the network.

\printbibliography

\end{document}